\documentclass{article}

\usepackage[final]{neurips_2019}

\usepackage[utf8]{inputenc}
\usepackage[T1]{fontenc}
\usepackage{hyperref}
\usepackage{url}
\usepackage{booktabs}
\usepackage{amsfonts}
\usepackage{nicefrac}
\usepackage{microtype}
\usepackage{graphicx}
\usepackage{xcolor}
\usepackage{lipsum}
\usepackage{float}
\usepackage[square,sort,comma,numbers]{natbib}

\title{
 Ensemble ALBERT on SQuAD 2.0\\
}

\author{
  Shilun Li \\
  Dept. of Mathematics \\
  Stanford University \\
  \texttt{shilun@stanford.edu} \\
  \And
  Renee Li \\
  Dept. of Computer Science \\
  Stanford University \\
  \texttt{reneeli@stanford.edu}\\
  \And
  Veronica Peng \\
  Dept. of Computer Science \\
  Stanford University \\
  \texttt{tpeng24@stanford.edu}
}

\begin{document}
\maketitle
\begin{abstract}
Machine question answering is an essential yet challenging task in natural language processing. Recently, Pre-trained Contextual Embeddings (PCE) models like Bidirectional Encoder Representations from Transformers (BERT) and A Lite BERT (ALBERT) have attracted lots of attention due to their great performance in a wide range of NLP tasks. In our CS224N Final Project, we utilized the fine-tuned ALBERT models and implemented combinations of additional layers (e.g. attention layer, RNN layer) on top of them to improve model performance on Stanford Question Answering Dataset (SQuAD 2.0). We implemented four different models with different layers on top of ALBERT-base model, and two other models based on ALBERT-xlarge and ALBERT-xxlarge. We compared their performance to our baseline model (\textbf{ALBERT-base-v2 + ALBERT-SQuAD-out}) with details. Our best-performing individual model is \textbf{ALBERT-xxlarge + ALBERT-SQuAD-out}, which achieved an F1 score of 88.435 on the dev set. Furthermore, we have implemented three different ensemble algorithms to boost overall performance. By passing in several best-performing models' results into our weighted voting ensemble algorithm, our final result \textbf{ranks $\bold{1^{st}}$ on the Stanford CS224N Test PCE SQuAD Leaderboard with $\bold{F1 = 90.123}$ (name: Libo).}

\end{abstract}

\section{Introduction}
For our project, we mainly focus on Stanford Question Answering Dataset (SQuAD 2.0) which approximate the real reading comprehension circumstances. Reading comprehension question-answering aims to answer questions given passages or documents. The goal of this project is to produce a question answering system that works well on SQuAD because models that perform well on SQuAD have been regarded as solid benchmarks to solve RC and QA problems. The QA task is always challenging since it not only requires a comprehensive understanding of natural languages, but also relies on the ability to do some further inference and reasoning. In recent years, many non-trivial NLP tasks, including those that have limited training data, have greatly benefited from pre-trained models. Academic research and papers have also shown that a large network is of high importance for achieving state-of-the-art performance. \citep{bert} \citep{radford} In fact, it has become common practice to pre-train large models and distill them down to smaller ones for real applications. \citep{sun} \citep{turc} Thus, our work mainly develops a model on top of the released ALBERT models. By replacing the linear ALBERT output layer with an encoder-decoder architecture or a highway network, we successfully implemented models that can deal with the SQuAD 2.0 dataset and the QA task quite well. We also tried building larger models with ALBERT-xlarge and ALBERT-xxlarge and were able to achieve satisfying results. We have also added ensemble after the model results come out to improve overall performance. Currently, our best single model has achieved an F1 score of 88.435 on the dev set, while the ensembled version achieved an F1 score of 89.734 on the dev set. Our leaderboard result is $F1 = 90.123$ on the test set.

\section{Related Works}
Language model pre-training has shown to be effective for improving many natural language processing tasks. Among different models, former google-released Bidirectional Encoder Representations from Transformers (BERT) \citep{bert} and the most recent google-released A Lite BERT for Self-Supervised Learning of Language Representations (ALBERT) \citep{albert_orig} are both empirically powerful models. We could potentially try more layers on top of BERT to improve its performance significantly; however, since we entered the PCE performance-oriented category of the CS224N SQuAD 2.0 Challenge, we have decided to mainly focus on finding literature about the leaner and more powerful ALBERT to achieve the best possible F1 and EM scores. ALBERT performs very well on a wide range of tasks, including the SQuAD 2.0 challenge, which is our task at hand. According to paper \citep{bert}, the pre-trained BERT representations can be fine-tuned to perform better in specific tasks. We wonder if we can improve the performance of ALBERT through additional architectures on top of ALBERT. Therefore, here we are going to build our own output network on top of pre-trained ALBERT. We will try building on top of both ALBERT with character embedding and ALBERT.

For data pre-processing, we relied heavily on SentencePiece tokenizer. SentencePiece is great for NMT because while existing subword segmentation tools assume that the input is pre-tokenized into word sequences, SentencePiece can train subword models directly from raw sentences, which helps us achieve better performance on QA tasks. \citep{sent_piece} For our baseline model, we decided to adapt ALBERT model with an ALBERT-SQuAD-out output layer so that we can clearly show whether a certain combination of layers on top of ALBERT can indeed improve its performance. ALBERT-SQuAD-out is essentially the linear output layer described in paper \citep{albert_orig} with some masking and other processing added, and it is described in more detail in Section \ref{albert_squad_out} below. We designed several modules on top of ALBERT as the task-specific output layers we want to test out to potentially improve model performance. We have a subsequent encoder-decoder architecture as post-processing to improve ALBERT model’s performance specifically on SQuAD 2.0. For our main encoder-decoder architecture, we have adapted RNN-based bi-directional long short-term memory layer (BiLSTM) \citep{lstm} and gated recurrent units (GRU) \citep{gru} as the encoder and decoder, which are commonly used in sequence to sequence translation task \citep{seq_to_seq}. As for the multi-layer state transitions, we used highway network \citep{highway} to adaptively copy or transform representations. We have also added more attention layers in transitions; more detailed explanation on how the attention layers turned out are offered in Experiments section as well. And for the final output layer, we compared the original ALBERT-SQuAD-out and BiDAF-out. The details are explained in the Approach section below.

Although our paper applied a variety of layers on top of the ALBERT model, it turns out that most of the layers do not improve the model performance of \textbf{ALBERT-base-v2 + ALBERT-SQuAD-out}. Previous papers exploring adding various layers over ALBERT or BERT have experienced similar setbacks. \citep{zhang_and_xu} \citep{haoshen_recommended_report} For example, Takeuchi, et al. (2018) has explored adding Highway layer, BiDAF, CNN, transformers, etc. over BERT, but the only model that ends up outperforming BERT + Linear is BERT + Highway. Most of the model fails to improve the performance of our baseline model because a lot of these layers force the model to learn non-existing correlations and thus add noises to the final predictions. As a result, it is within expectation that additional layers don't necessarily improve the \textbf{ALBERT-base-v2 + ALBERT-SQuAD-out} performance. More explanation to why adding more layers doesn't necessarily improve model performance is included in the Experiments and Analysis sections below.

\section{Approach}

\subsection{Baseline and oracle}

\textbf{Baseline: ALBERT-base-v2 + ALBERT-SQuAD-out} uses the fine-tuned ALBERT (albert-base-v2 \citep{albert}) with a ALBERT-SQuAD-out layer on top. We use \textbf{ALBERT-base-v2 + ALBERT-SQuAD-out} as a baseline because we want to understand how much information can be extracted using the embedding provided by ALBERT. The baseline reaches $EM = 79.91$ and $F1 = 82.03$.

\textbf{Oracle: Human performance} is used as a proxy for Bayes error by convention. We used the human performance given in the SQuAD dataset to be our oracle: $EM = 86.83$, $F1 = 89.45$. \citep{SQuAD}




    



\subsection{Our Model: ALBERT + various layers} \label{modules}
In this section, we discuss model architectures we have implemented or we plan to implement. We used fine-tuned ALBERT (albert-base-v2 \citep{albert}) to generate contextualized token embeddings. On top of ALBERT, we try 4 different combinations of layers: ALBERT-SQuAD-out, Highway + ALBERT-SQuAD-out, and bi-LSTM Encoder + Attention + bi-LSTM Decoder + BiDAF-out, GRU Encoder + Highway + GRU Decoder + BiDAF-out, and GRU Encoder + Attention + Self-Attention + GRU Decoder + BiDAF-out. The detailed explanation and visualization of the architecture are shown below. After trying these combinations, we will put the results from all models through 3 ensemble algorithms, as explained in detail in Section \ref{ensemble}.

\begin{center}
    \includegraphics[width=0.8\textwidth]{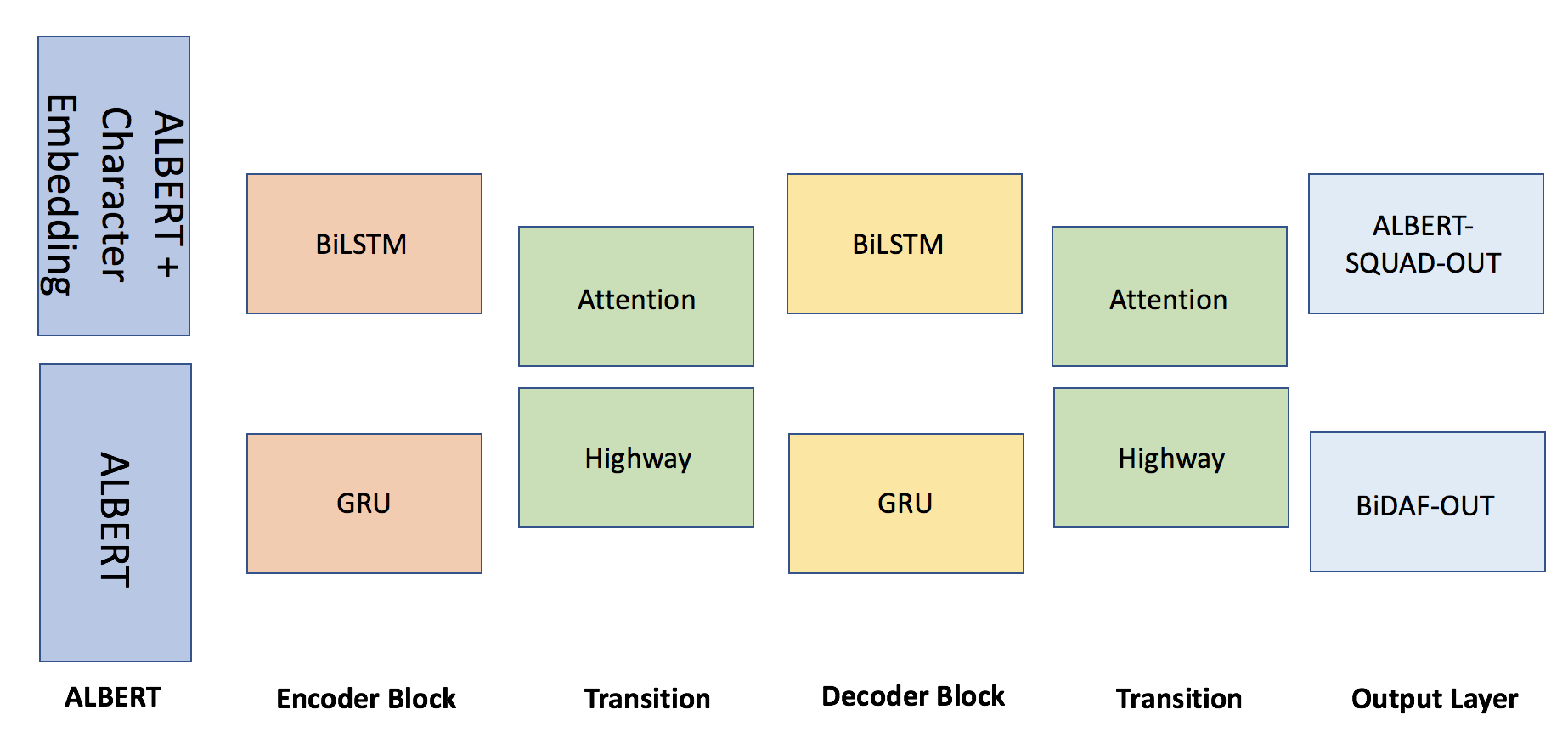}

Figure 1. Schema of Model Architecture
\end{center}

As to coding, we significantly modified the code for data pre-processing and the eval file from huggingface/transformer. \citep{pre_processing} The baseline code provided by CS224N is based on word representation, but our code is based on token representation, so we have made significant changes to every file and method. \textbf{The ALBERT model is borrowed from huggingface code base, but all the layers on top of ALBERT are implemented by us.} We also wrote all of our ensemble algorithms from scratch.

\subsubsection{Embedding Layer}
We used two varieties of embedding layers. The first type of embedding directly comes from the ALBERT models, and is labeled with the ALBERT model name in the following literature. (e.g. \textit{ALBERT-xlarge} if the embedding is generated using ALBERT-xlarge). The second type of embedding is a combination of pre-trained character embedding given in the starter code and the ALBERT token embeddings. It will be labeled as \textit{(ALBERT model name + char)} in the following literature (\textit{(ALBERT-base-v2 + char)} if the embedding is generated using ALBERT-base-v2). 

To generate the combined embedding, we first add a weighted average attention to the original token embedding, which is proved to be useful in detecting unanswerable questions by previous work. \citep{weighted-average-attention} The weighted average attention is generated by $softmax(\textbf{E}\textbf{W})$ where E is the token embedding and W is the learnable weights. For the pre-trained character embedding, we first use CNN to pool information from the original embedding, and then add the weighted average attention of the character embedding to itself. As a final step, we concatenate the processed token embedding and character embedding. Then, we further refine the concatenated embedding with a 2-layer highway encoder (See Section \ref{highway-network} for more details on highway encoder). Adding the character embedding to the original embedding is supposed to provide further information about word meaning, and should be especially helpful when the input contains misspelled words, typos or new words.

\subsubsection{Encoder and Decoder Blocks}
We explored several encoders and decoders discussed in other NLP works and have decided to explore the following two:

\textbf{Bidirectional Long Short-Term Memory (BiLSTM) Layer Encoder/Decoder:} Long Short-Term Memory (LSTM) is an RNN architecture aimed to solve vanishing gradients problem. On each time step $t$, we have a hidden state $h_t$ and a cell state $c_t$. The cell can store long-term information, and the LSTM can erase, write and read information from the cell controlled by three gates (forget gate $f_t$, input gate $i_t$ and output gate $o_t$). By adding an LSTM encode/decoder on top of the ALBERT models, we can integrate temporal dependencies between time-steps of the output tokenized sequence better. \citep{lstm}

\textbf{Gated Recurrent Units (GRU) Encoder/Decoder:} Gated Recurrent Units [4] is a simpler alter- native to the LSTM. And on each time step t, we have input x(t) and hidden state h(t) (no cell state), and we used two gates (update gate u(t) and reset gate r(t)) to control the states. \citep{gru}

\subsubsection{Self-Attention}
We add a self-attention right after the attention layer. Self-attention, sometimes called intra-attention, is an attention mechanism relating different positions of a single sequence in order to compute a representation of the sequence. \citep{attention} Self-attention has been successfully applied to a variety of tasks including reading comprehension and abstractive summarization. For some NLP tasks, it is suggested to add attention mechanisms on top of the LSTM layer to bring in an extra source of information to guide the extraction of sentence embedding. In our model, we have implemented a self-attention layer discussed in paper \citep{attention}, and it allows each position of the output token to attend to all positions up to and including that position. As part of our implementation and experiments, we have used the basic \textbf{dot-product attention}. Within the BiDAF model, we also used \textbf{sub-attention} on top of regular attention.

\subsubsection{Highway Network} \label{highway-network}
Highway network is a novel architecture that enables the optimization of networks with virtually arbitrary depth. By applying a gating mechanism, a neural network can have paths along which information can flow across several layers without attenuation. As shown in lecture and past assignment, $W_{proj}$, $W_{gate}$, $b_{gate}$ and $b_{proj}$ are learnable parameters. We choose a highway network to train multi-layer state transitions in our recurrent neural networks. Instead of traditional neural layers, it can allow the network to adaptively copy or transform representations. So it can help to refine the tokenized sequence. \citep{highway} \citep{modules} 


\subsection{Output Layer}
For the output layer, we tried the original ALBERT-SQuAD-out output layer and the QA output layer from Bi-Directional Attention Flow (BiDAF-Out).

\subsubsection{ALBERT-SQuAD-out} \label{albert_squad_out}
We built upon the simple linear output layer that converts the dimension of the output sequence from ($batch\_size$, $seq\_len$, $hiddenstate$) to ($batch\_size$, $seq\_len$, 2). And we split it to get the start and end logits. Then, we masked out the logits that are not the original context. Finally, we compute the cross-entropy loss with the start and end position vectors.

\subsubsection{BiDAF-out}
To replace the linear output layer in default ALBERT model, we add a QA output layer named \textit{BiDAF-out} adapted from BiDAF model. We get the start logits by summing the linear transformations of outputs of attention layer and the decoder. To get the end logits, we first feed the output of the decoder to RNN to get a representation for the end token. Then we produced the end logits by summing the linear transformations of the attention layer's outputs and the representation for the end token. As a final step, we masked out the logits for the question to make sure we only select tokens in the context paragraph.

\subsection{Ensemble} \label{ensemble}
As we have mentioned above, we put the results from different combinations of our models into our three ensemble algorithms. We first fed the results from our best performing models (four trained \textbf{ALBERT-xxlarge + ALBERT-SQuAD-out}) into a \textbf{mean logits} ensemble; mean logits basically adds the logits together which is basically finding the joint probability of the models. Another ensemble strategy we tried is \textbf{weighted voting}: the ensemble gives the model's F1 score as weight to the model's best start and end words token prediction, sums the weights for each prediction, and select the prediction with the highest weight. In the first experiment with \textbf{weighted voting}, we used 4 \textbf{ALBERT-xxlarge + ALBERT-SQuAD-out} models with different initialization for ensemble. In the second experiment with \textbf{weighted voting}, we used the 4 \textbf{ALBERT-xxlarge + ALBERT-SQuAD-out} models mentioned above together with mean logits ensemble mentioned above to produce the \textbf{weighted voting with mean logits} ensemble method. Details about performance and analysis can be found in Section \ref{results} and Section \ref{results_analysis}.

\section{Experiments}

\subsection{Data}
\textbf{Dataset:} We used Stanford Question Answering Dataset (SQuAD 2.0) \citep{dataset} to train and evaluate our models. Each entry in this dataset includes: 1) a question, 2) 3 human answers to the question, and 3) a context paragraph where the answer lies. The context paragraphs are excerpts from Wikipedia. The questions and answers were crowdsourced using Amazon Mechanical Turk. The dataset is split into train (129,941 questions), dev (6078 questions) and test (5915 questions) sets. For each set, roughly half of the questions cannot be answered using the provided context paragraph. However, if the question is answerable, the answer is a continuous span of text taken directly from the paragraph. With this dataset, our task is to predict whether the paragraph contains the answer to the question, and if yes, where the answers are located. 

\textbf{Data pre-processing: Transform word indices into token indices:} The original dataset provides the word indices of the start and end word of the answers. However, since ALBERT is based on subword tokens, we relabelled the word indices of the context paragraph as token indices. An example of tokenization and a detailed walk-through is provided in the Appendix (Section \ref{appendix_1}).

\textbf{Data pre-processing: Divide context paragraphs that are too long:} In order to make better use of the limited GPU memory, we limit the max number of tokens in a context paragraph (max\_seq\_length) to 384 for ALBERT-base-v2 \footnote{max\_seq\_length = 280 for ALBERT-xlarge and ALBERT-xxlarge because large models require more GPU memory, so more of the paragraphs will exceed max\_seq\_length in those two cases.}. However, 10\% of the paragraphs exceeds the max\_seq\_length. For these paragraphs, we divided them into new paragraphs with at most max\_seq\_length tokens. Each paragraph has the original question as their new question. For the chunk with the answer in it, the answer is the original answer. For the other chunks, the answer is N/A. An example of how we processed a long context paragraph is given in the Appendix (Section \ref{appendix_1}).

    
    
\subsection{Evaluation method} \label{evaluation_method}
    
    

\textbf{EM} is 1 when the model prediction matches the ground truth exactly, and 0 otherwise. 

\textbf{F1} is the harmonic mean of precision and recall ($F1 = \frac{2 \times prediction \times recall}{precision + recall}$). Precision refers to what percentage of words in predicted answer is in the ground truth answer, and recall refers to what percentage of words in the ground truth answer is in predicted answer. 

Since each question has three human-provided answers, we took the maximum F1 and EM scores across all three answers for that question. See Section \ref{Evaulation-score} for examples of EM and F1 score calculations.


    

\subsection{Experimental details} \label{experiment}
For \textbf{ALBERT-base-v2 + ALBERT-SQuAD-out}, we used $batch\_size = 7$, and for all other models involving \textbf{ALBERT-base-v2}, we used $batch\_size = 5$. For all \textbf{ALBERT-base-v2} models, we had $learning\_rate = 3 \times 10^{-5}$, $epochs = 3$, $max\_seq\_length = 384$, $doc\_stride = 128$, $dropout\_rate = 0.2$. It took about 250,000 to 300,000 steps for the \textbf{ALBERT-base-v2} models to converge, where a single step refers to training on a single example. Since there are about 129,941 examples in the train set, we trained the \textbf{ALBERT-base-v2} models only for 3 epochs to prevent the model from overfitting. We chose the batch size and max\_seq\_length specifically to use as much GPU memory as possible in training without going out of memory. With this set of hyperparameter, our 4 8GB Tesla M60 GPUs uses 88.24\% of the memory while training the \textbf{ALBERT-base-v2 + ALBERT-SQuAD-out} model. 

For \textbf{ALBERT-xlarge + ALBERT-SQuAD-out} and \textbf{ALBERT-xxlarge + ALBERT-SQuAD-out}, we used $batch\_size = 1$, $max\_seq\_length = 280$, $doc\_stride = 128$, $dropout\_rate = 0.2$. The learning rate for \textbf{ALBERT-xlarge + ALBERT-SQuAD-out} is $1 \times 10^{-5}$, and the learning rate for \textbf{ALBERT-xxlarge + ALBERT-SQuAD-out} is $8 \times 10^{-6}$. It took around 100,000 to 200,000 steps for the \textbf{ALBERT-xlarge + ALBERT-SQuAD-out} models to converge, and around 50,000 to 100,000 steps for the \textbf{ALBERT-xxlarge + ALBERT-SQuAD-out} models to converge, so we ran the \textbf{ALBERT-xLarge + ALBERT-SQuAD-out} for 2 epochs, and the \textbf{ALBERT-xxLarge + ALBERT-SQuAD-out} for about 1 epochs.

    
    
    

\subsection{Results} \label{results}

\begin{table}[H]
\label{tab:baseline_model_single}
\centering
\caption{Single model performance on dev set}
 \resizebox{\columnwidth}{!}{
\begin{tabular}{@{}lll@{}}
\toprule
                                                                                            & \textbf{EM}    & \textbf{F1} \\ \midrule
\textbf{BiDAF}                                                                              & 57.99          & 61.23       \\
\textbf{ALBERT-base-v2 + ALBERT-SQuAD-out (Baseline)}                                       & 79.91          & 82.03       \\
\textbf{ALBERT-base-v2 + Highway + ALBERT-SQuAD-out}                                        & 79.88          & 82.19       \\ 
\textbf{(ALBERT-base-v2 + char) + bi-LSTM Encoder + Attention + bi-LSTM Decoder + BiDAF-out}& 79.17          & 81.57       \\
\textbf{(ALBERT-base-v2 + char) + GRU Encoder + Highway + GRU Decoder + BiDAF-out}                          & 78.22          & 80.65       \\
\textbf{(ALBERT-base-v2 + char) + GRU Encoder + Attention + Self-Attention + GRU Decoder + BiDAF-out}       & 78.63          & 81.04       \\
\textbf{ALBERT-xlarge + ALBERT-SQuAD-out}                                                   & 83.71          & 86.05       \\
\textbf{ALBERT-xxlarge + ALBERT-SQuAD-out}                                                  & \textbf{85.57} & \textbf{88.43} \\ \bottomrule
\end{tabular}}
\label{results1}
\end{table}

\begin{table}[H]
\label{tab:baseline_model_ensemble}
\centering
\caption{Ensemble model performance on dev set}
\begin{tabular}{@{}lll@{}}
\toprule
                                                        & \textbf{EM}    & \textbf{F1} \\ \midrule
\textbf{Mean logits}                                    & 86.84          & 89.62       \\
\textbf{Weighted voting}                                & \textbf{87.02} & \textbf{89.73}       \\
\textbf{Weighted voting with mean logits}               & 86.94          & 89.69       \\ \bottomrule
\end{tabular}
\label{results2}
\end{table}
%



We are currently (at the submission time of this paper) ranked \textbf{number one on the PCE test leaderboard with $F1 = 90.123$ and $EM = 87.303$.}

From the results, we can see that adding more layers on top of ALBERT-base-v2 model (with char embedding) did not bring about significant improvement to the performance of the model: the only model that offered minor improvement according to F1 score was the \textbf{ALBERT-base-v2 + Highway + ALBERT-SQuAD-out} model. This lack of improvement in performance came as a surprise because we expected that adding more layers such as attention should improve the model performance. We offer detailed analysis of why this may have happened in Section \ref{results_analysis}.

We can also see from the table that larger ALBERT model is the key to boosting performance. As shown in Table \ref{results1}, \textbf{ALBERT-xxlarge + ALBERT-SQuAD-out} has the best performance, with $EM = 85.571$ and $F1 = 88.435$. While \textbf{ALBERT-xlarge + ALBERT-SQuAD-out} doesn't perform as well as the xxlarge model, it also also performs better than our baseline (\textbf{ALBERT-base-v2 + ALBERT-SQuAD-out}. We offer analysis on why this happens in Section \ref{results_analysis} as well.

According to Table \ref{results2}, all the ensemble strategies we used tend to improve the performance by about $1.5$ in terms of F1 score. Although all the ensemble strategies have similar performance, the best ensemble model is weighted voting, with a F1 score of 89.73. However, surprisingly, adding the ensembled mean logits model to the weighted voting ensemble doesn't improve the performance probably because the mean logits doesn't include information other than what the 4 single models can provide.

\section{Analysis}
\subsection{Comparing Models and Results} \label{results_analysis}
According to Table 2, all ALBERT models outperform the BiDAF model, indicating that the token embedding generated by ALBERT provides more contextual information than using BiDAF alone. Interestingly, adding additional information from char embedding doesn't improve the performance. An explanation is that the token embedding from ALBERT already captures a lot of sub-word structures, so adding char embedding is providing repetitive information and it won't improve the performance. Similarly, adding more complicated layers including highway and attention on top of ALBERT also doesn't make the model performance significantly better than the baseline \textbf{ALBERT-base-v2 + ALBERT-SQuAD-out}. An explanation is that since our ALBERT models take the context and question together as input and the transformer in ALBERT has multi-head attention, the token embeddings generated by ALBERT itself already incorporate better attention between context and question than the extra attention layer we add on top. However, the model performance does start to increase when we use larger ALBERT models including ALBERT-xlarge and ALBERT-xxlarge. The performance improves probably because the larger ALBERT models have larger hidden layer size, so they can better capture attention than the smaller ALBERT-base-v2 model. 

\subsection{Error Analysis on Selected Examples}


\subsubsection{Example study 1}
\begin{itemize}
    \item \textbf{Question:} Private networks were connected via gateways for what reason? 
    \item \textbf{Context Paragraph:} The private networks were often connected via gateways to the public network to reach locations not on the private network.
    \item \textbf{Human Answers 1, 2, and 3 are N/A} 
    \item \textbf{Answer by ALBERT-xxlarge + ALBERT-SQuAD-out:} to reach locations not on the private network
    \item \textbf{The error:} Humans think the question has no answer, but the model predicts an answer.
    \item \textbf{Potential causes:} Wrong answers are given by humans.
    \item \textbf{Potential fix:} Correct the wrong human answers in the train, dev and test set to improve the accuracy. However, the very fact that the model is able to predict an answer even when the human is wrong indicates that the model is able to draw general patterns learnt from the train set to be resistant to errors.
\end{itemize}

\subsubsection{Example study 2}
\begin{itemize}
    \item \textbf{Question:} What car is licensed by the FSO Car Factory and built in Egypt?
    \item \textbf{Context Paragraph:} The FSO Car Factory was established in 1951. A number of vehicles have been assembled there over the decades, including the Warszawa, Syrena, Fiat 125p (under license from Fiat, later renamed FSO 125p when the license expired) and the Polonez. The last two models listed were also sent abroad and assembled in a number of other countries, including Egypt and Colombia. 
    \item \textbf{Human Answers 1 and 2:} Polonez
    \item \textbf{Human Answer 3:} 125p
    \item \textbf{Answer by ALBERT-xxlarge + ALBERT-SQuAD-out:} 
    \item \textbf{The error:} The model fails to predict an answer when there is one.
    \item \textbf{Potential causes:} The keywords in the question (e.g. FSOO, Egypt license) are far away from each other in the context paragraph. Thus, the likelihood of the answer is more sparsely distributed among the words in the three sentence, making the probability of "no answer" even higher than the actual answers.
    \item \textbf{Potential fix:} To preserve the memory for longer, we could potentially add skip connections to our RNN encoders and decoders when training.
\end{itemize}

\subsubsection{Example study 3}
\begin{itemize}
    \item \textbf{Question:} How many residential dorms house upper class, sophomore, Jr, and Sr students?
    \item \textbf{Context Paragraph:} Sophomore, junior, and senior undergraduates live in twelve residential Houses, nine of which are south of Harvard Yard along or near the Charles River. 
    \item \textbf{Human Answer 1:} twelve residential Houses
    \item \textbf{Human Answers 2 and 3:} twelve
    \item \textbf{Answer by ALBERT-xxlarge + ALBERT-SQuAD-out:} 
    \item \textbf{The error:} The model fails to predict an answer when there is one.
    \item \textbf{Potential causes:} The keywords in the question (e.g. Jr, and Sr) are not present in the context. Instead, their synonyms (e.g. junior, and senior) exist in the context. The model is having problem building connections between the synonyms. The problem with synonyms appears in this example probably because Jr and Sr are uncommon abbreviations of the full word, that it is hard to build the connection even with the insight into sub-word structure provided by the token embeddings.
    \item \textbf{Potential fix:} Build a dictionary of abbreviation and its full word, so that the model can borrow insight from the dictionary when it encounters abbreviation with uncommon sub-word structures.
\end{itemize}

\subsubsection{Example study 4}
\begin{itemize}
    \item \textbf{Question:} In what year did Protestant rule in Montpellier effectively collapse?
    \item \textbf{Context Paragraph:} Even before the Edict of Alès (1629), Protestant rule was dead and the ville de sûreté was no more.[citation needed]
    \item \textbf{Human Answers 1, 2, and 3 are N/A} 
    \item \textbf{Answer by ALBERT-xxlarge + ALBERT-SQuAD-out:} (1629),
    \item \textbf{The error:} The model provides an answer when there is no answer to the question.
    \item \textbf{Potential causes:} The model understands the question is asking for a year, but it doesn't capture the nuance in the meaning of the question. The question asks for when the Protestant rule in Monpellier effectively collapse, and the context says the the Protest rule was dead even before 1629. Thus, we know the Protestant rule was dead before 1629, but we don't know necessarily when.
    \item \textbf{Potential fix:} In order for the model to better capture the nuance of the question, we could build better attention between the question and the context by trying bi-linear or tri-linear attention.
\end{itemize}

\section{Conclusion} \label{conclusion}
The main goal of the project is to search for models that make predictions about answers as accurately as possible. In this paper, we designed several task-specific architectures on top of the ALBERT model according to insights gained from other networks. As mentioned in Section \ref{experiment}, we have implemented 4 combinations of layers (encoder and decoder layer, self attention layer, highway network, and output layer) mentioned in Section \ref{modules} on top of ALBERT-base-v2; beyond that, we have also implemented \textbf{ALBERT-xlarge + ALBERT-SQuAD-out} and \textbf{ALBERT-xxlarge + ALBERT-SQuAD-out}. Finally, we have tried three different ensemble methods with different combinations of model results.

By comparing their performance to \textbf{ALBERT-base-v2 + ALBERT-SQuAD-out} baseline model and doing a deep analysis, we finally implemented our best model with an \textbf{ALBERT-xxlarge + ALBERT-SQuAD-out}. With ensemble techniques, our model achieves an F1 score of 89.734 on the Dev Set and 90.123 on the Test Set. We can further improve the performance of our model by fine-tuning the parameters in each layer. A limitation is that we had to truncate the sequences because we could not fit them into the GPU otherwise. Based on the error analysis, having longer sequences could have helped improve the performance of our models; with longer sequences, the sentences that used to be split into two sequences could be fit into one without losing the context, and our models can generate more accurate answers that way.

For potential future work, we could consider adding some extra layers inside of ALBERT to further improve its performance. We could also perform a more comprehensive error analysis to understand what types of questions the model performs poorly on, and improve performance based on the understanding; for example, questions starting with 'Why', 'When', 'Who' and 'How', and 'Yes/No' questions.

\section{Appendix} \label{appendix}
\subsection{Data Pre-Processing Examples}
\subsubsection{Example for Data pre-processing:  Transform word indices into token indices} \label{appendix_1}
\begin{itemize}
    \item \textbf{Question: } In which month was Obama born?
    \item \textbf{Context paragraph: } Obama was born in August.
    \item \textbf{Tokenized paragraph: } O ba ma\_was\_born\_in\_Au gust.
    \item \textbf{Token indices: } [[0, 5], [0, 5], [0, 5], [6, 9], [10, 14], [15, 17], [18, 24], [18, 24]]
    \item \textbf{Possible model output: } [6, 6]
    \item \textbf{Answer: } August.
\end{itemize}
Each token has a corresponding tuple (the index of the first character for the word it belongs to, the index of the last character for the word it belongs to). As seen in the example above, \textit{Obama} is divided into three tokens \textit{O}, \textit{ba} and \textit{ma}. For the first token \textit{O}, its corresponding index is [0, 5] since it belongs to the word \textit{Obama} whose initial character \textit{O} is indexed 0 and last character \textit{a} is indexed 5 in the context paragraph. The model predicts [6, 6] for the answer to the question which means the start and the end token are both the \textit{6th} token \textit{Au}. The token \textit{Au}'s index is [18, 24], so the answer to the question is \textit{August}, which spans from the \textit{18'th} to the \textit{24'th} characters in the original context paragraph.

\subsubsection{Example for Data pre-processing: Divide context paragraphs that are too long} \label{appendix_2}
\begin{itemize}
    \item \textbf{Max\_seq\_length:} 5
    \item \textbf{Original data entry:}
    \begin{itemize}
        \item \textbf{Question:} How old is Jay?
        \item \textbf{Tokenized context:} \_jay \_is \_12 \_years \_old . \_he \_lives \_in \_flo mo .
        \item \textbf{Answer:} 12
    \end{itemize}
    \item \textbf{New data entry 1:}
    \begin{itemize}
        \item \textbf{Question:} How old is Jay?
        \item \textbf{Tokenized context:} \_jay \_is \_12 \_years \_old .
        \item \textbf{Answer:} 12
    \end{itemize}
    \item \textbf{New data entry 2:}
    \begin{itemize}
        \item \textbf{Question:} How old is Jay?
        \item \textbf{Tokenized context:} \_he \_lives \_in \_flo mo.
        \item \textbf{Answer:} N/A
    \end{itemize}
\end{itemize}

\subsection{Evaluation Method Exmaples}
\label{Evaulation-score}
\subsubsection{EM score example}
\begin{table} [H]
\centering
\caption{examples to illustrate EM score}
\begin{tabular}{lll}
\toprule
\textbf{Ground Truth}& \textbf{Prediction} & \textbf{EM} \\ \midrule
Albert Einstein      & Einstein            & 0       \\
Albert Einstein      & Albert Einstein     & 1       \\ \bottomrule
\end{tabular}
\end{table}

\subsubsection{F1 score example}
\begin{itemize}
    \item \textbf{Predicted answer:} Einstein
    \item \textbf{Ground truth answer:} Albert Einstein
    \item \textbf{Precision:} 100\%
    \item \textbf{Recall:} 50\%
    \item \textbf{F1}: 
    $$F1 = \frac{2 \times prediction \times recall}{precision + recall} = \frac{2 \times 100 \times 50}{100 + 50} = 66.67\%$$
\end{itemize}

\bibliographystyle{unsrt}
\bibliography{references}

\begin{thebibliography}{10}

\bibitem{bert}
Jacob Devlin, Ming-Wei Chang, Kenton Lee, and Kristina Toutanova.
\newblock Bert: Pre-training of deep bidirectional transformers for language
  understanding.
\newblock {\em arXiv:1810.04805}, 2018.

\bibitem{radford}
Alec Radford, Jeffrey Wu, Rewon Child, David Luan, Dario Amodei, and Ilya
  Sutskever.
\newblock Language models are unsupervised multitask learners.
\newblock {\em OpenAI Blog, 1(8)}, 2019.

\bibitem{sun}
Siqi Sun, Cheng Yu, Zhe Gan, and Jingjing Liu.
\newblock Patient knowledge distillation for bert model compression.
\newblock {\em arXiv:1908.09355}, 2019.

\bibitem{turc}
Ilulia Turc, Ming-Wei Chang, Kenton Lee, and Kristina Toutanova.
\newblock Well-read students learn better: The impact of student initialization
  on knowledge distillation.
\newblock {\em arXiv:1908.08962}, 2019.

\bibitem{albert_orig}
Zhenzhong Lan, Mingda Chen, Sebastian Goodman, Kevin Gimpel, Piyush Sharma, and
  Radu Soricut.
\newblock Albert: A lite bert for self-supervised learning of language
  representations.
\newblock {\em arXiv:1909.11942v6}, 2019.

\bibitem{sent_piece}
Taku Kudo and John Richardson.
\newblock Sentencepiece: A simple and language independent subword tokenizer
  and detokenizer for neural text processing.
\newblock 2018.

\bibitem{lstm}
Jianpeng Cheng, Li~Dong, and Mirella Lapata.
\newblock Long short-term memory-networks for machine reading.
\newblock {\em arXiv:1601.06733}, 2016.

\bibitem{gru}
Kyunghyun Cho, Bark Van~Merriënboer, Caglar Gulcehre, Dzmitry Bahdanau, Fethi
  Bougares, Holger Schwenk, and Yoshua Bengio.
\newblock Learning phrase representations using rnn encoder- decoder for
  statistical machine translation.
\newblock {\em arXiv:1406.1078}, 2014.

\bibitem{seq_to_seq}
Sean Robertson.
\newblock Nlp from scratch: Translation with a sequence to sequence network and
  attention, 2017.
\newblock Available at \url{https://pytorch.org/tutorials/intermediate/seq2seq_
  translation_tutorial.html, 2018.}

\bibitem{highway}
Rupesh Srivastava, Klaus Greff, and Jürgen Schmidhuber.
\newblock Highway networks.
\newblock {\em arXiv:1505.00387}, 2015.

\bibitem{zhang_and_xu}
Yuwen Zhang and Zhaozhuo Xu.
\newblock Bert for question answering on squad 2.0.
\newblock 2019.

\bibitem{haoshen_recommended_report}
Danny Takeuchi and Kevin Tran.
\newblock Improving squad 2.0 performance using bert + x.
\newblock 2018.

\bibitem{albert}
Transformers: State-of-the-art natural language processing for tensorflow 2.0
  and pytorch, 2018.
\newblock Available at \url{https://github.com/huggingface/transformers}.

\bibitem{SQuAD}
Squad2.0 the stanford question answering dataset, 2020.
\newblock Available at \url{https://rajpurkar.github.io/SQuAD-explorer/}.

\bibitem{pre_processing}
Transformers: State-of-the-art natural language processing for tensorflow 2.0
  and pytorch - run\_squad.py, 2018.
\newblock Available at
  \url{https://github.com/huggingface/transformers/blob/master/examples/run_squad.py}.

\bibitem{weighted-average-attention}
Mingchen Li, Gendong Zhang, and Zixuan Zhou.
\newblock Improved bidaf with self-attention.
\newblock 2018.
\newblock Available at
  \url{https://web.stanford.edu/class/cs224n/reports/default/15734514.pdf}.

\bibitem{attention}
Ashish Vaswani, Noam Shazeer, Niki Parmar, Jakob Uszkoreit, Llion Jones,
  Aidan~N. Gomez, Łukasz Kaiser, and Illia Polosukhin.
\newblock Attention is all you need.
\newblock {\em Adv. Neural Inform. Process. Syst. (NIPS), 2017, pp.
  6000–6010.}, 2017.

\bibitem{modules}
Yuwen Zhang and Zhaozhuo Xu.
\newblock Bert for question answering on squad 2.0, 2018.
\newblock Available at
  \url{https://web.stanford.edu/class/archive/cs/cs224n/cs224n.1194/reports/default/15848021.pdf}.

\bibitem{dataset}
Pranav Rajpurkar, Jian Zhang, Konstantin Lopyrev, and Percy Liang.
\newblock Squad: 100,000+ questions for machine comprehension of text.
\newblock {\em arXiv:1606.05250}, 2016.

\end{thebibliography}

\end{document}